\documentclass[letterpaper, 10 pt, conference, dvipsnames]{ieeeconf}  
\usepackage{multirow}
\usepackage{makecell}
\usepackage{amsmath}
\usepackage{amssymb}
\usepackage{booktabs}
\usepackage{graphicx}
\usepackage{tabularx}
\usepackage{ragged2e}
\usepackage{colortbl}
\usepackage{algorithm}
\usepackage{algorithmicx}
\usepackage{algpseudocode}
\usepackage{caption}    
\usepackage{subcaption}  
\usepackage[table]{xcolor}
\usepackage[dvipsnames]{xcolor}
\usepackage{pifont}
\let\labelindent\relax
\usepackage{enumitem}
\usepackage{tikz}
\usetikzlibrary{positioning, shapes, arrows.meta, calc}
\usepackage{csquotes}
\usepackage[
    colorlinks=true,
    urlcolor=black,
    linkcolor=black,
    citecolor=black
]{hyperref}

\IEEEoverridecommandlockouts                              

\title{\LARGE \bf
More than A Point: Capturing Uncertainty with Adaptive Affordance Heatmaps for Spatial Grounding in Robotic Tasks
}

\author{Xinyu Shao$^{1,2,*}$, Yanzhe Tang$^{1,2,*}$,Pengwei Xie$^{2}$, Kaiwen Zhou$^{2}$,  \\
Yuzheng Zhuang$^{2}$, Xingyue Quan$^{2}$, and Jianye Hao$^{2}$, Long Zeng$^{1}$, and Xiu Li$^{1}$ 
\thanks{*Equal contribution.}%
\thanks{$^{1}$Shenzhen International Graduate School, Tsinghua University, China.}%
\thanks{$^{2}$Noah’s Ark Lab, Huawei, China.}%
}

\begin{document}

\maketitle
\thispagestyle{empty}
\pagestyle{empty}

\begin{abstract}

Many language-guided robotic systems rely on collapsing spatial reasoning into discrete points, making them brittle to perceptual noise and semantic ambiguity. To address this challenge, we propose RoboMAP, a framework that represents spatial targets as continuous, adaptive affordance heatmaps. This dense representation captures the uncertainty in spatial grounding and provides richer information for downstream policies, thereby significantly enhancing task success and interpretability. RoboMAP surpasses the previous state-of-the-art on a majority of grounding benchmarks with up to a 50x speed improvement, and achieves an 82\% success rate in real-world manipulation. Across extensive simulated and physical experiments, it demonstrates robust performance and shows strong zero-shot generalization to navigation. More details and videos can be found at \href{https://robo-map.github.io/}{robo-map.github.io}.
\end{abstract}

\section{INTRODUCTION}

Vision-Language Models have recently shown impressive capabilities in perception and reasoning, naturally motivating their use in language-guided robotics~\cite{zhou2023language,xiao2025robot,li2025generative,huang2025roboground,zitkovich2023rt}. A prominent paradigm is the hierarchical approach, where a VLM
serves as a high-level planner~\cite{yuan2024robopoint,team2025robobrain,song2025robospatial,yang2025bridging}. This process, known as \textbf{spatial grounding}, translates instructions and observations into a mid-level representation to guide a low-level controller. An effective intermediate representation is key to bridging perception and action, enabling precise and versatile robotic behaviors. 

However, the choice of this intermediate representation is critical and remains a key limitation. Most existing methods distill complex language and visual information into deterministic keypoints~\cite{yuan2024robopoint}, bounding boxes or visual traces~
\cite{yuan2025seeing}. This creates a severe information bottleneck. By discarding all spatial uncertainty, it makes the system brittle. 

This challenge is most pronounced when the robot's goal is defined by a spatial relationship (e.g., \emph{near the chair}) rather than by targeting a specific, named object (e.g., \emph{the red chair}). Collapsing such an inherently ambiguous region into a single keypoint fails to capture its spatial extent. This makes the resulting action fragile and hard to interpret, especially in cluttered or dynamic environments.

\begin{figure}[t]
    \centering
    \includegraphics[width=\linewidth,height=5.5cm]{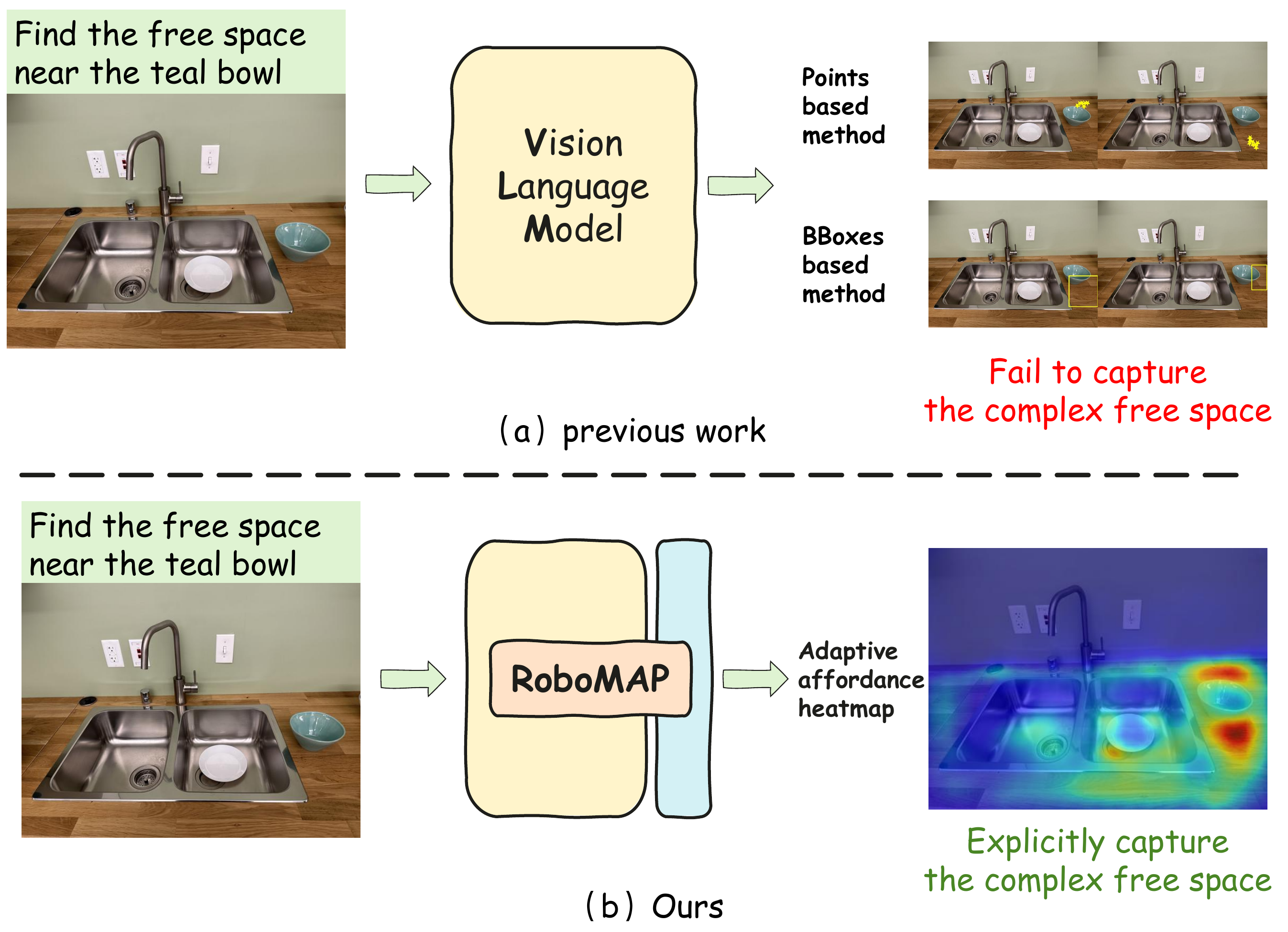}
    \caption{Given an ambiguous instruction like \emph{Find the free space near the teal bowl}, prior works (a) that rely on discrete points or bounding boxes fail to capture the complex, non-rectangular nature of the goal region. In contrast, our RoboMAP (b) generates a dense affordance heatmap that accurately represents the entire continuous distribution of suitable locations, successfully grounding the complex spatial concept.}
    \label{fig:teaser}
    \vspace{-2mm}
\end{figure}

To address this brittleness, we argue that robust language-guided robotics requires more than a point. We introduce \textbf{RoboMAP} (\textit{\textbf{M}ore than \textbf{A} \textbf{P}oint}), a framework that generates our core intermediate representation: the \textbf{adaptive affordance heatmap}. This heatmap models the entire workspace as a continuous probability distribution, quantifying the suitability of each region for a given language-conditioned task. Figure~\ref{fig:teaser} provides a clear example of its effectiveness: while discrete representations like points or bounding boxes (a) are brittle, our dense heatmap (b) successfully captures the inherent spatial uncertainty in complex instructions.

To generate these dense heatmaps in a scalable and data-efficient manner, we develop a pre-training pipeline that converts diverse, sparse annotations (keypoints, bounding boxes, and trajectories) into dense supervisory signals. The resulting adaptive affordance heatmap endows the system with two critical advantages: robustness, by providing multiple high-probability locations for the robot to attempt, and safety, as the robot can refuse to act if the peak of the generated heatmap falls below a predefined confidence threshold.

Our contributions are three-fold:
\begin{itemize}
\item \textbf{Adaptive Heatmaps for Robust Grounding}: We introduce a framework that leverages an adaptive affordance heatmap for robust spatial grounding in ambiguous, object-free regions where traditional points-based or bounding boxes-based representations fall short.

\item \textbf{Unifying Sparse Labels into Heatmaps}: A pre-training pipeline that unifies diverse and sparse annotations (e.g., keypoints, boxes, trajectories) by synthesizing them into a consistent adaptive affordance heatmap. This data-efficient approach enables cross-domain learning.

\item \textbf{Zero-Shot Generalization}: We validate RoboMAP's effectiveness through extensive experiments in both simulation and the real world. We deploy the system on diverse platforms, demonstrating successful zero-shot transfer to a range of complex tasks, such as pick-and-place, industrial sorting, and navigation.
\end{itemize}

\section{Related Work}

\subsection{VLM-driven Language-Guided Robotics}

Recent progress in language-guided robotics is largely driven by the powerful reasoning capabilities of Vision-Language Models (VLMs). Two dominant paradigms have emerged: end-to-end Vision-Language-Action (VLA) models and hierarchical approaches that use VLMs for high-level planning. End-to-end models, like RT-2~\cite{zitkovich2023rt}, OpenVLA~\cite{kim2024openvla} and $\pi_0$~\cite{black2024pi0visionlanguageactionflowmodel}, directly map language commands to low-level executable actions. Hierarchical approaches~\cite{ahn2022can,ji2025robobrain,yuan2024robopoint,song2025robospatial}, which we detail in Sec.~\ref{mid-level_representation}, use VLMs to generate an intermediate representation to guide a low-level controller.

Despite their architectural differences, both paradigms share a common limitation in grounding instructions that refer to object-free regions, e.g., \emph{place the block to the left of the bowl}. End-to-end VLA models struggle to directly map abstract spatial relations to precise action vectors, a process that is often brittle and data-intensive. Hierarchical approaches are constrained by their object-centric intermediate representations, which ground an object like \textit{the bowl} but discard crucial information about the surrounding free space. This reveals a shared need for a representation capable of explicitly modeling continuous space and spatial relations, which motivates our work.

\subsection{Intermediate representation-based Spatial Grounding}
\label{mid-level_representation}

In the hierarchical paradigm, an intermediate spatial representation bridges high-level planning and low-level control. The design of this representation is critical, as it directly dictates the system's capabilities and limitations when grounding complex spatial instructions.

\textbf{Point-based and BBox-based methods} represent the target as discrete 2D coordinates. While some works predict multiple keypoints~\cite{yuan2024robopoint}, and others enhance the reasoning process~\cite{liu2025spatialcot, song2025robospatial}, they ultimately output a few sparse points or a bounding box. This discrete representation is fundamentally limited as it collapses a potentially large, valid region into an overconfident prediction. By discarding information about the region's shape and the distribution of affordances, this approach makes the system brittle.  

\textbf{Trajectory-based methods} offer more detailed guidance by predicting an explicit path for the robot to follow, such as a visual trace or a post-contact trajectory~\cite{yuan2025seeing, a02504}. While this is useful for complex actions like pushing, the approach is still limited to a single, deterministic path. Critically, this static representation cannot adapt to the task's ambiguity, failing to broaden the possibilities for vague instructions or narrow them for precise ones~\cite{li2025generative}.

\textbf{Heatmap-based methods}, while used for tasks like grasp planning~\cite{chen2024efficient} or localizing interaction points~\cite{xu2024bridgevla}, have two core limitations. They are typically \textbf{non-adaptive}, acting as fixed-shape pointers unable to model task ambiguity. Furthermore, they are not used as a \textbf{unified representation} for learning from diverse 2D internet-scale data, often requiring specialized 3D simulations.

Our framework, RoboMAP, overcomes these limitations with its \emph{adaptive affordance heatmaps}. Unlike prior representations that are discrete, deterministic, or non-adaptive, ours is a truly probabilistic and dynamic distribution. It adapts its shape to match the task's ambiguity and serves as a unified format for learning from varied 2D data sources. This enables RoboMAP to provide a richer and more flexible signal for the robot, leading to more robust and interpretable execution. 

\section{Problem Formulation}

Given an RGB image $I \in \mathbb{R}^{H \times W \times 3}$ and a language instruction $x$, we aim to learn a unified representation for spatial grounding across diverse robotic tasks. This representation must handle both precise targets (e.g., \textit{pick up the screw}) and high-uncertainty ambiguous regions (e.g., \textit{near the window} or \textit{any of the empty drawers}).

We formulate this as predicting a single, dense adaptive affordance heatmap, $\hat{M} \in [0,1]^{H \times W}$. Formally, we train a Vision-Language Model $f_{\theta}$ that learns the mapping:
$$
f_{\theta}(I, x) \rightarrow \hat{M}.
$$
The predicted heatmap $\hat{M}$ is a continuous probability distribution that dynamically adapts its shape to reflect the instruction's uncertainty, forming sharp peaks for precise targets and broad distributions for vague or disjoint regions.

To train this model, a key component of our approach is to generate a ground-truth heatmap $M^*$ as the supervision signal for each training sample. Since dense heatmap annotations are not readily available, we synthesize these labels from various common, sparse annotation types, including Keypoints ($P^*$), Bounding boxes ($B^*$), and Trajectories ($T^*$), as detailed in Section~\ref{sec:heatmap_synthesis}.

\begin{figure*}[htbp]
    \centering
    \includegraphics[width=\textwidth]{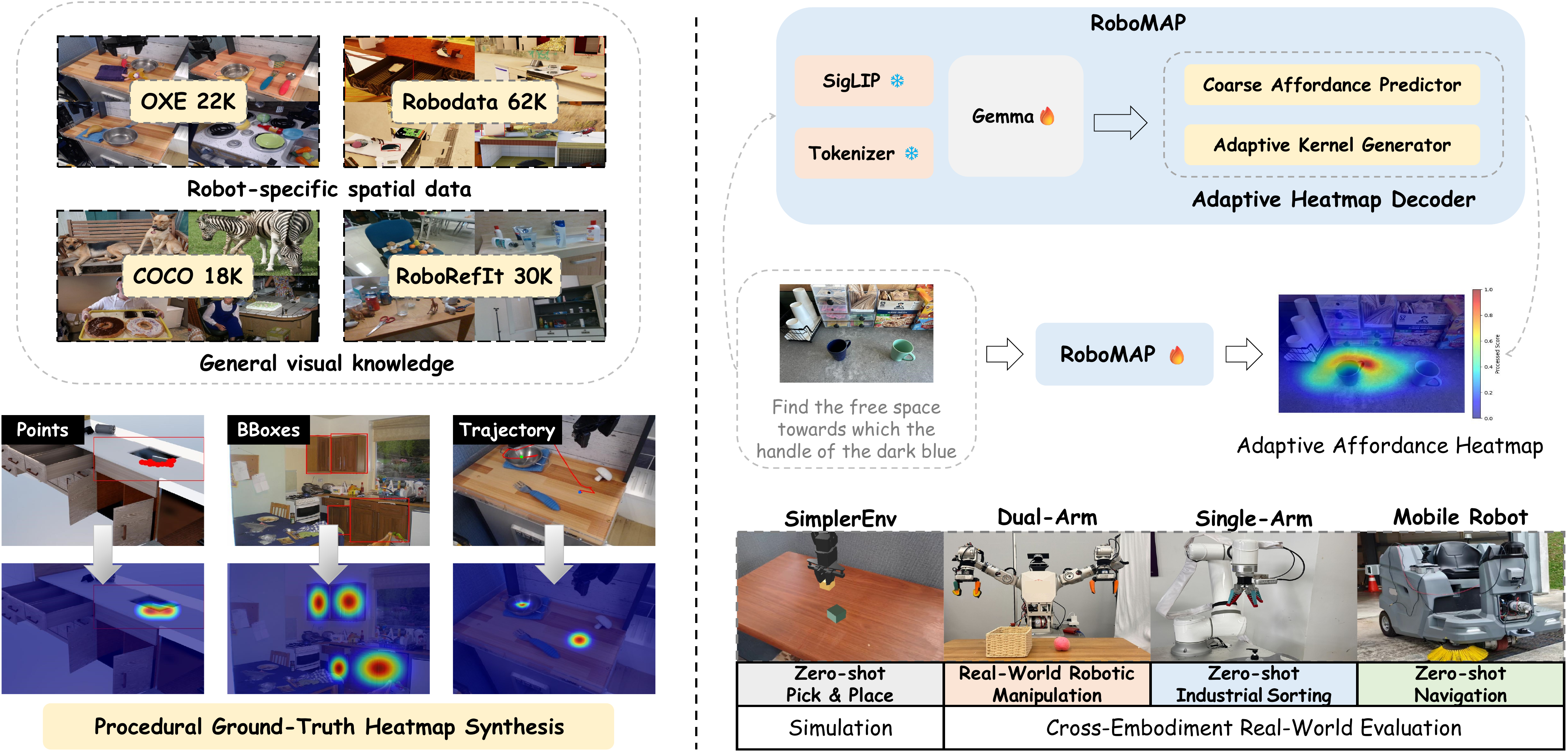}
    \caption{Overall Architecture of the RoboMAP Framework.}
    \label{fig:overall_architecture}
\end{figure*}

\section{Methodology: The RoboMAP Framework}

RoboMAP is a framework for spatial grounding in robotic tasks. It combines a vision-language backbone with our proposed Adaptive Heatmap Decoder (AHD) to produce adaptive affordance heatmaps. To leverage heterogeneous supervision, we introduce a procedural heatmap synthesis strategy and a heatmap-based training objective. The overall pipeline is shown in Fig.~\ref{fig:overall_architecture}.

\subsection{Overall Architecture}
RoboMAP builds upon a vision-language backbone to extract semantically grounded, spatially aligned features for robotic tasks. We adopt \textbf{PaliGemma}~\cite{beyer2024paligemma} for its ability to produce language-conditioned visual tokens that remain consistent with the input image grid after multi-modal fusion. Given an RGB image $I$ and a natural language instruction $x$, the backbone jointly encodes both modalities and outputs a sequence of multi-modal tokens. A subset of these tokens, corresponding to visual content, is reshaped into a low-resolution feature map $F_{\text{low}}$ (e.g., $16\times16$). This map captures the spatial structure and serves as the input to our Adaptive Heatmap Decoder, which generates the final high-resolution affordance heatmap $\hat{M}$.

\subsection{Adaptive Heatmap Decoder (AHD)}

\begin{figure*}[htbp]
    \centering
    \includegraphics[width=0.95\textwidth]{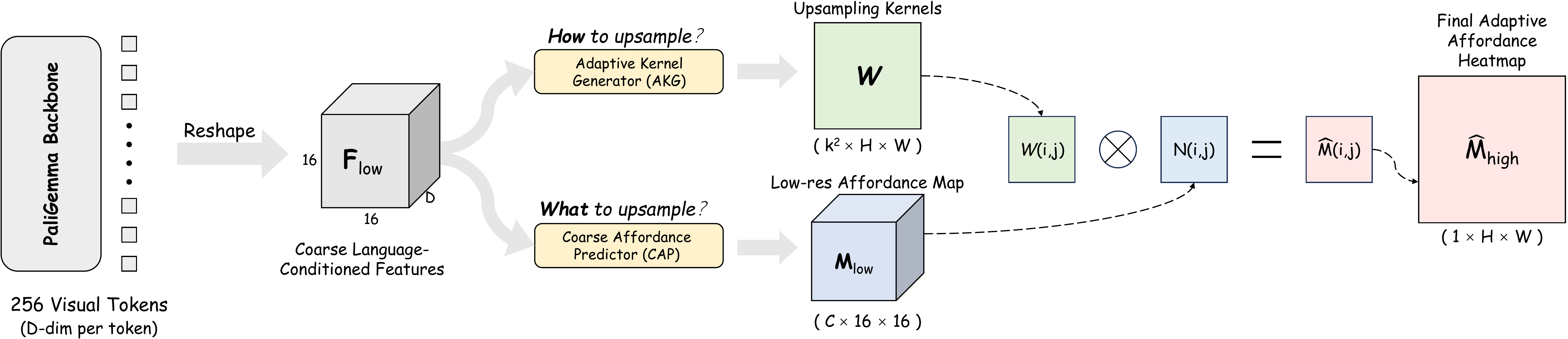}  
    \caption{The architecture of Adaptive Heatmap Decoder (AHD). A \textbf{PaliGemma Backbone} processes the image and instruction into visual tokens. These are \textbf{reshaped} into a language-conditioned feature grid $F_{\text{low}}$, which is then fed into our AHD. The AHD uses two branches (AKG and CAP) to compute the final adaptive affordance heatmap.}
    \label{fig:ahd_architecture}
\end{figure*}

Generating high-resolution heatmaps for the variable shapes of robotic affordances is a key challenge, as common upsampling methods are fundamentally content-agnostic. For instance, Bilinear interpolation ignores feature semantics, Deconvolution~\cite{zeiler2014visualizing} applies a single, computationally expensive global kernel, and PixelShuffle~\cite{shi2016real} rearranges pixels according to a fixed, content-agnostic rule. To overcome these limitations, our AHD adopts a content-aware upsampling mechanism inspired by CARAFE~\cite{wang2019carafe}. As illustrated in Figure~\ref{fig:ahd_architecture}, our AHD implements this by decomposing the task into two parts: an \textbf{Adaptive Kernel Generator (AKG)} that learns \emph{how} to upsample, and a \textbf{Coarse Affordance Predictor (CAP)} that provides the content of \emph{what} to upsample.

The two branches take the coarse feature map $F_{\text{low}}$ as input. The CAP first derives a preliminary low-resolution affordance estimate $M_{\text{low}}$. In parallel, the AKG outputs a unique $k \times k$ kernel $W(i,j)$ for each target pixel $(i,j)$, which is normalized by a Softmax function. The final heatmap value $\hat{M}(i,j)$ is then computed by applying the predicted kernel to its corresponding neighborhood in the coarse map:

\begin{equation}
\label{eq:ahd}
\hat{M}(i,j) = \sum_{p,q \in \mathcal{N}_k} W(i,j)_{p,q} \cdot M_{\text{low}}(i'+p, j'+q),
\end{equation}
where $(i', j')$ are the source coordinates in the low-resolution map corresponding to the target pixel $(i,j)$, and $\mathcal{N}_k$ defines the $k \times k$ neighborhood, with $p$ and $q$ being the integer offsets.

This learnable convex upsampling allows the AHD to adapt its behavior to the instruction's ambiguity. For precise targets, the predicted kernels are sharp and focused; for vague regions, they become broader and smoother. This results in high-fidelity, shape-adaptive heatmaps that accurately capture spatial uncertainty and improve downstream task success rates across diverse scenarios. 

\subsection{Procedural Ground-Truth Heatmap Synthesis}
\label{sec:heatmap_synthesis}
A key component of RoboMAP is the procedural generation of high-quality supervision signals from diverse raw annotations (e.g., keypoints, bounding boxes, trajectories), as illustrated in Fig~\ref{fig:overall_architecture}. For each training sample, we generate a ground-truth adaptive affordance heatmap $M^*$ that encodes the target's spatial properties.

\subsubsection{Aggregated Point Supervision}
When supervision is provided as one or more discrete points $\{p_i^*\}_{i=1}^N$, we generate the heatmap by aggregating Gaussian distributions centered at these points:

\begin{equation}
M^*(u) = \max_{i} \exp\left(-\frac{\|u - p_i^*\|^2}{2\sigma^2}\right),
\label{eq:point_heatmap}
\end{equation}
where $u$ is a pixel coordinate. For a single-point ($N=1$), we use a small, fixed variance $\sigma$ to create a sharp peak, promoting precise localization. For multiple points, a \emph{max} operation, instead of summation, is used to prevent blurring, thus preserving clear peaks that help the model distinguish between close targets.

\subsubsection{Shape-Aware Bounding Box Supervision}
For supervision provided as an axis-aligned bounding box $B^*$, we generate an elliptical Gaussian heatmap that matches the box's aspect ratio. This is achieved by decoupling the Gaussian into two components along the x and y axes:

\begin{equation}
M^*(u) = \exp\left( -\left( \frac{(u_x - c_x)^2}{2\sigma_x^2} + \frac{(u_y - c_y)^2}{2\sigma_y^2} \right) \right),
\label{eq:bbox_heatmap_axis_aligned}
\end{equation}
where $u=(u_x, u_y)$ is a pixel coordinate, $c=(c_x, c_y)$ is the center of the bounding box. The standard deviations $\sigma_x$ and $\sigma_y$ are set to be proportional to the box's width and height, respectively. This creates a heatmap that is elongated or compressed according to the object's shape, providing richer geometric cues than a simple circular Gaussian.

\subsubsection{Mask-based Trajectory Supervision}
\label{sec:trajectory_supervision}

To leverage large-scale robotics datasets like OXE~\cite{o2024open}, which provide supervision as action trajectories incompatible with standard VLM training, we introduce a pipeline to synthesize ground-truth heatmaps. This process converts the implicit spatial goals within trajectories into a dense supervision signal, directly enabling VLMs to learn the robust and generalizable spatial reasoning required for robotic tasks.

Our pipeline first uses GPT-4o~\cite{hurst2024gpt} to identify the manipulated object from the instruction. We then track this object using Grounding DINO~\cite{liu2024grounding} and SAM~\cite{kirillov2023segment} to extract a goal mask from the trajectory's final frame, which is then re-projected to the initial frame. This binary mask is converted into a heatmap using a distance transform with a Gaussian kernel (Eq.~\ref{eq:trajectory_heatmap}). This entire process results in our new dataset of (image, instruction, heatmap) triplets, which we name \textbf{OXE-MAP}, providing rich supervision for our model.

\begin{equation}
M^*(u) = \exp\left(-\frac{\min_{m \in M_{\text{goal}}}\|u - m\|^2}{2\sigma^2}\right).
\label{eq:trajectory_heatmap}
\end{equation}

\subsection{Training Objective}

We formulate the training as a dense prediction task. The model is optimized by minimizing the pixel-wise Binary Cross-Entropy (BCE) loss between the predicted heatmap $\hat{M}$ and the procedurally generated ground-truth heatmap $M^*$. This objective effectively encourages the network to produce sharp, high-confidence peaks for precise targets while permitting broader distributions for ambiguous goals, all without requiring additional hyperparameters.

\section{Implementation Details}
RoboMAP uses PaliGemma~\cite{beyer2024paligemma} as the backbone, with our lightweight Adaptive Heatmap Decoder appended for adaptive heatmap prediction. This decoder, consisting of two branches (a Coarse Affordance Predictor and an Adaptive Kernel Generator), contains only 1.6M parameters, a negligible amount compared to the backbone.

All input images are resized to $224 \times 224$ pixels. During training, we apply only photometric data augmentations (e.g., brightness, contrast, and color jitter) without geometric transformations such as rotation or flipping. Geometric augmentations are avoided because they would distort the spatial correspondence between the image and the ground-truth heatmap, potentially introducing ambiguity and hindering the model's ability to learn precise language-conditioned spatial grounding. The training set is a combination of Robodata (62K)~\cite{yuan2024robopoint}, COCO (22K)~\cite{lin2014microsoft}, OXE-MAP (18K)~\cite{o2024open} and RoboRefIt (30K)~\cite{zhou2025roborefer}, for a total of \textbf{132K} samples.

We finetune the model for 2 epochs. The vision encoder and token embeddings remain frozen. We use AdamW optimizer~\cite{loshchilov2017decoupled} with $\beta_1=0.9$, $\beta_2=0.999$, and weight decay of 0.1. The learning rate warms up to a peak of $3\times10^{-5}$ over the first 400 steps, followed by cosine decay. Training is performed with a global batch size of 384 on four 48G GPUs used for all experiments, using \texttt{BF16} mixed precision~\cite{kalamkar2019study}. The full process takes approximately 20 hours. All experiments are implemented in PyTorch~\cite{paszke2019pytorch}.

\begin{table*}[t]
\centering
\caption{Performance and efficiency comparison on robotic grounding benchmarks.}
\label{tab:main_results}
\footnotesize
\begin{tabularx}{\textwidth}{l cc *{4}{>{\centering\arraybackslash}X} c} 
\toprule
\multirow{2}{*}[-1.5ex]{\textbf{Model}} & 
\multirow{2}{*}[-1.5ex]{\textbf{Size}} & 
\multirow{2}{*}[-1.5ex]{\textbf{\begin{tabular}[c]{@{}c@{}} Fine-tuning \\ Method \end{tabular}}} &
\multicolumn{4}{c}{\textbf{Accuracy (\%) $\uparrow$}} &
\multirow{2}{*}[-1.5ex]{\textbf{\begin{tabular}[c]{@{}c@{}}Inference \\ Time (s) $\downarrow$\end{tabular}}} \\
\cmidrule(lr){4-7}
& & & \textbf{Where2place} & \textbf{RoboRefIt} & \textbf{RefSpatial} & \textbf{VABench-Point} & \\
\midrule
\multicolumn{8}{l}{\textit{General-Purpose VLMs}} \\
\midrule
GPT-4o~\cite{hurst2024gpt} & - & - & 23.04 & 15.28 & 9.12 & 16.59 & - \\
Gemini-2.5-pro~\cite{comanici2025gemini} & - & - & 52.95 & 49.50 & 29.24 & 21.71 & - \\
Qwen2.5VL~\cite{bai2025qwen25vltechnicalreport} & 72B & - & 37.15 & 78.50 & 20.85 & 23.30 & - \\
GLM-4.5v~\cite{glm2024chatglm} & 106B & - & 29.89 & 74.50 & 11.19 & 29.23 & - \\
\midrule
\multicolumn{8}{l}{\textit{Specialized Robotic Grounding VLMs}} \\
\midrule
RoboPoint~\cite{yuan2024robopoint} & 13B & SFT & 46.80 & 49.80 & 16.07 & 19.09 & 3.10 \\
RoboRefer~\cite{zhou2025roborefer} & 2B & SFT & 66.00 & 72.80& 33.77 & 24.67 & 1.54 \\
RoboBrain2.0~\cite{team2025robobrain} & 3B & - & 54.86 & 54.42 & 35.81 & 8.08 & 0.67 \\
RoboBrain2.0~\cite{team2025robobrain} & 7B &  SFT + RFT & 63.59 & 47.55 & 32.50 & 12.70 & 10.41 \\
FSD~\cite{yuan2025seeing} & 13B & SFT & 45.81 & 56.73 & 14.90 & 61.82 & 15.68 \\
Embodied-R1~\cite{yuan2025embodied} & 3B & SFT + RFT & \underline{69.50} & \underline{85.58} & \textbf{38.46} & \underline{66.00} & 2.18 \\
BridgeVLA-pretrain~\cite{xu2024bridgevla} & 3B & SFT & 15.00 & 26.25 & 11.50 & 10.50 & \underline{0.06} \\
\midrule
\rowcolor{blue!10}
\textbf{RoboMAP} & 3B & SFT & \textbf{73.00} & \textbf{88.73} & \underline{36.50} & \textbf{70.00} & \multicolumn{1}{c}{\textbf{0.04}} \\
\bottomrule
\end{tabularx}
\end{table*}

\section{Experiments}
\label{sec:experiments}

In this section, we conduct a series of experiments to rigorously evaluate the proposed RoboMAP framework. Our evaluation aims to answer three primary questions: 

\textbf{(Q1)} How does RoboMAP perform against state-of-the-art and baseline methods in spatial grounding tasks? 

\textbf{(Q2)} What is the contribution of each core component? 

\textbf{(Q3)} How well does RoboMAP generalize to novel robotic tasks and platforms, enabling effective zero-shot generalization in both simulation and the real world?

To answer these questions, we structure our experiments as follows: Section~\ref{sec:grounding} addresses Q1, providing a rigorous quantitative and qualitative evaluation against state-of-the-art baselines on four grounding benchmarks. Section~\ref{sec:ablation} then addresses Q2, conducting a thorough ablation study to pinpoint the contribution of each key design choice. Finally, Section~\ref{sec:deployment} tackles Q3, demonstrating RoboMAP's powerful generalization by successfully testing its zero-shot transfer to both simulated and diverse real-world platforms.

\subsection{Benchmark Evaluation on Spatial Grounding (Q1)}
\label{sec:grounding}

\paragraph{Benchmarks}
\label{sec:benchmark}

We evaluate RoboMAP on four benchmarks, each targeting a key aspect of spatial grounding: RoboRefIt for disambiguating objects in clutter; Where2place for grounding instructions in object-free regions; RefSpatial for multi-step localization and placement; and VABench for translating instructions into actionable regions. These benchmarks collectively serve to evaluate the model's core capability in spatial grounding, primarily measured through localization accuracy.

\paragraph{Baselines}
We compare RoboMAP against two main categories of baselines to ensure a comprehensive evaluation. The first category includes leading \emph{general-purpose VLMs}, such as GPT-4o~\cite{hurst2024gpt}, Gemini-2.5-pro~\cite{comanici2025gemini}, Qwen2.5VL~\cite{bai2025qwen25vltechnicalreport}, and GLM-4.5v~\cite{glm2024chatglm}. To test their intrinsic grounding capabilities, these models are evaluated in a strict \emph{zero-shot} setting, prompted to directly output target coordinates based on the visual and language input. The second category comprises state-of-the-art models specifically designed for \emph{robotic spatial grounding}. This includes top-performing \emph{points-based} methods like RoboRefer~\cite{zhou2025roborefer}\footnote{We evaluate the publicly available SFT-trained version of RoboRefer, as its RFT-trained counterpart has not been open-sourced.}, RoboBrain2.0~\cite{team2025robobrain}, RoboPoint~\cite{yuan2024robopoint}, FSD~\cite{yuan2025seeing}, and Embodied-R1~\cite{yuan2025embodied}. This comparison directly contrasts RoboMAP's dense, uncertainty-aware heatmap approach against these established discrete prediction methods. Additionally, we include a quantitative and qualitative comparison with the prior heatmap-based VLM, BridgeVLA-pretrain~\cite{xu2024bridgevla}.

\begin{figure*}[t]
    \centering
    \includegraphics[width=\linewidth]{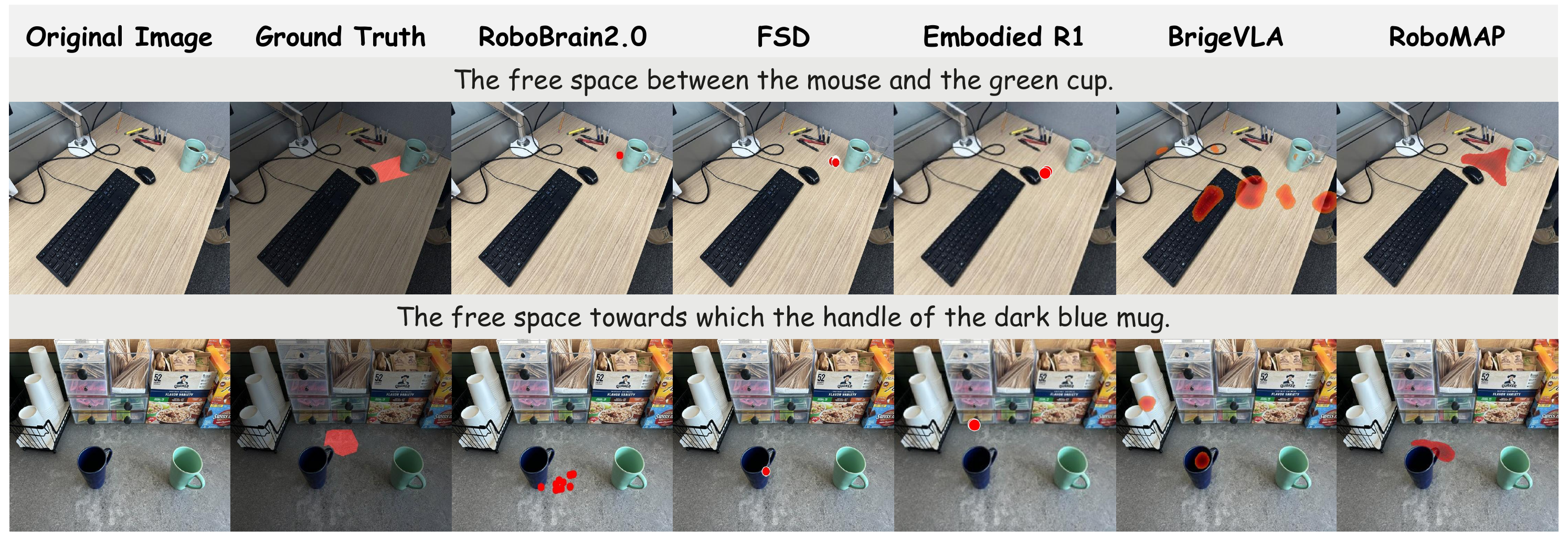}
    \caption{Qualitative comparison on challenging spatial grounding instructions. The visualizations highlight RoboMAP's ability to generate coherent heatmaps for ambiguous regions. This contrasts with the scattered points or diffuse heatmaps produced by baseline methods in the same scenarios.}
    \label{fig:qualitative_vis}
\end{figure*}

\paragraph{Quantitative Evaluation}

We evaluate RoboMAP against a comprehensive suite of VLMs in Table~\ref{tab:main_results}. While RoboMAP's core advantage is its rich spatial heatmaps, current benchmarks are limited to point-based evaluation, requiring us to \textbf{simplify our output for a fair comparison}. Specifically, we extract only the single highest-confidence point from our heatmaps, meaning the following results are achieved with our model's capabilities partially constrained.

Even \textbf{under this constrained evaluation}, RoboMAP demonstrates an outstanding balance of performance and efficiency, as shown in the table. It establishes new state-of-the-art results on three of the four key benchmarks.

The source of this strong performance lies in our core design. RoboMAP's ability to capture spatial uncertainty makes it exceptionally effective on benchmarks requiring reasoning about ambiguous regions. Its significant lead on RoboRefIt further shows its strength in disambiguating objects in clutter. RoboMAP remains highly competitive on the multi-step RefSpatial benchmark, despite not being explicitly designed for sequential reasoning. Its architecture is optimized to predict an instruction's final outcome at once.

Perhaps most notably, RoboMAP achieves its top-ranked performance with a significantly simpler and more efficient methodology. Unlike the next-best competitor, \textbf{Embodied-R1}, which requires both Reinforced Fine-tuning (RFT) and an explicit reasoning stage, our model relies solely on a standard Supervised Fine-tuning (SFT) stage and a Direct inference pass. Critically, this results in an inference time of just \textbf{0.04 seconds}, making it \textbf{over 50 times faster} than Embodied-R1. This combination of state-of-the-art accuracy, architectural simplicity, and exceptional efficiency validates RoboMAP as a powerful and practical foundation for spatial grounding in robotic tasks.

\paragraph{Qualitative Analysis}

We present representative visualizations comparing RoboMAP with baseline models in Fig~\ref{fig:qualitative_vis}. These examples illustrate how RoboMAP captures broader, uncertain regions. This enables more robust and interpretable spatial grounding. In contrast, baselines relying on discrete points often fail to account for spatial uncertainty, especially in object-free or cluttered areas.

The specific advantages of our model are evident. When tasked to identify object-free spaces, such as \textit{between the mouse and the green cup} (top row), the baseline models exhibit significant limitations, each in distinct ways. The point-based models exhibit two primary error patterns: some make clear localization errors by incorrectly targeting the surrounding objects, while others place scattered points in the correct region but fail to represent it as a coherent region. The other heatmap-based baseline, \textbf{BridgeVLA-pretrain}, also demonstrates a complete inability to ground instructions in object-free areas. In contrast, RoboMAP successfully generates a well-defined intermediate representation, forming a heatmap that is both focused on the target area and expressive of its spatial extent and uncertainty.

\subsection{Ablation Studies (Q2)}
\label{sec:ablation}
We conduct extensive ablation studies to validate two core components of our design: (i) the effectiveness of our proposed heatmap decoder, and (ii) the impact of the hybrid data composition strategy.

\begin{table}[t]
\centering
\small
\caption{Ablation study on the heatmap decoder architecture.}
\label{tab:decoder_ablation}
\resizebox{\columnwidth}{!}{%
\begin{tabular}{l cccc}
    \toprule
    \textbf{Decoder Architecture} & \textbf{Where2place} & \textbf{RoboRefIt} & \textbf{RefSpatial} & \textbf{VABench} \\
    \midrule
        Deconvolution~\cite{zeiler2014visualizing} & \underline{64.00\%} & \underline{81.00\%} & 23.00\% & \underline{69.00\%} \\
    PixelShuffle~\cite{shi2016real} & 59.00\% & 80.00\% & \underline{29.00\%} & 68.00\% \\
    Bilinear & \underline{64.00\%} & 80.70\% & 26.39\% & 68.00\% \\
    \midrule
    \rowcolor{blue!10}
    Ours (AHD) & \textbf{73.00}\% & \textbf{88.73\%} & \textbf{36.50\%} & \textbf{70.00\%} \\
    \bottomrule
\end{tabular}%
}
\end{table}

\begin{table}[t]
    \centering
    \small
    \caption{Ablation study on the composition of training data.}
    \label{tab:data_ablation}
    \resizebox{\columnwidth}{!}{%
    \begin{tabular}{l cccc}
        \toprule
        \textbf{Training Data} & \textbf{Where2place} & \textbf{RoboRefIt} & \textbf{RefSpatial} & \textbf{VABench} \\
        \midrule
        w/ RoboRefIt only & 5.00\% & \underline{79.50\%} & 15.20\% & 5.50\% \\
        w/ COCO only & 15.20\% & 67.80\% & 12.00\% & 6.50\% \\
        w/ OXE-MAP only & \underline{48.00\%} & 56.80\% & \underline{20.20\%} & \underline{65.50\%} \\
        w/ RoboData only & 40.00\% & 54.50\% & 13.40\% & 11.00\% \\
        \midrule
        \rowcolor{blue!10}
        Ours (Full Data Mix) & \textbf{73.00}\% & \textbf{88.73\%} & \textbf{36.50\%} & \textbf{70.00\%}  \\
        \bottomrule
    \end{tabular}%
    }
\end{table}

\paragraph{Analysis of Decoder Architecture}


As shown in Table~\ref{tab:decoder_ablation}, our Adaptive Heatmap Decoder (AHD) establishes a new state-of-the-art on all benchmarks with gains of up to 9\%. This superior performance comes from its ability to generate adaptive heatmaps that intelligently adjust to scene content, overcoming a key limitation of static, content-agnostic decoders.

\paragraph{Analysis of Training Data Composition}

The results in Table~\ref{tab:data_ablation} validate our hybrid data strategy, as the full data mix achieves the top performance across all benchmarks. This demonstrates that unifying diverse datasets is highly effective for balancing general grounding with robotics-specific understanding, creating a model that avoids the specialization trade-offs inherent in single-source training.

\subsection{Zero-Shot Generalization and Deployment (Q3)}
\label{sec:deployment}

To answer Q3, we evaluate RoboMAP's ability to generate effective spatial representations for diverse robotic embodiments in a zero-shot setting, without any task-specific fine-tuning. We test its capabilities on both manipulation scenarios and a completely new task domain: navigation.

\subsubsection{Embodied Performance on Manipulation Tasks}

\paragraph{Performance in Simulation}
We first evaluate our method in SimplerEnv~\cite{li2024evaluatingrealworldrobotmanipulation}, a scalable simulator for manipulation policies. In these tests, RoboMAP generates affordance heatmaps in a zero-shot manner to guide a motion planner. As shown in Table~\ref{tab:simplerenv_widowx_succ}, our method achieves a \textbf{60.5\%} success rate, establishing a new state-of-the-art among VLM-based methods. It outperforms prior leading works such as Embodied-R1 (56.2\%) and SoFar (53.8\%). This demonstrates RoboMAP's strong foundational ability to ground instructions in complex, interactive scenarios.

\begin{figure*}[t]
    \centering
    {
        \captionsetup{type=table}
        \footnotesize
        \caption{Evaluation of Real-World Robotic Manipulation Tasks.}
        \label{tab:realworld_evaluation}
        \begin{tabularx}{\textwidth}{l *{5}{>{\centering\arraybackslash}X} cc}
            \toprule
            \multirow{3}{*}{\textbf{Method}} & \multicolumn{5}{c}{\textbf{Successes (out of 10 trials)}} & \multirow{3}{*}{\textbf{\begin{tabular}[c]{@{}c@{}}Success \\ Rate\end{tabular}}} & \multirow{3}{*}{\textbf{\begin{tabular}[c]{@{}c@{}}Inference \\ Speed (s)\end{tabular}}} \\
            \cmidrule(lr){2-6}
            & \textbf{\begin{tabular}[c]{@{}c@{}}Place [A] \\ on [B]\end{tabular}}
            & \textbf{\begin{tabular}[c]{@{}c@{}}Move [A] \\ beside [B]\end{tabular}}
            & \textbf{\begin{tabular}[c]{@{}c@{}}Move [Specified A] \\ to [B]\end{tabular}}
            & \textbf{\begin{tabular}[c]{@{}c@{}}Place [A] \\ relative to [B]\end{tabular}}
            & \textbf{\begin{tabular}[c]{@{}c@{}}Place [A] into the \\ empty area in [B]\end{tabular}}
            & & \\
            \midrule
            FSD & 2/10 & 3/10 & 4/10 & 4/10 & 1/10 & 28\% & 15.68 \\
            RoboBrain2.0-3B & 5/10 & \underline{7/10} & \underline{7/10} & \textbf{7/10} & \underline{7/10} & \underline{66\%} & \underline{0.45} \\
            RoboBrain2.0-7B & \underline{8/10} & \underline{7/10} & 6/10 & 5/10 & 6/10 & 64\% & 8.07 \\
            \midrule
            \rowcolor{blue!10}
            \textbf{RoboMAP} & \textbf{9/10} & \textbf{9/10} & \textbf{8/10} & \underline{6/10} & \textbf{9/10} & \textbf{82\%} & \textbf{0.04} \\
            \bottomrule
        \end{tabularx}
    }
    \\[1.5em]
    \begin{minipage}[t]{0.49\textwidth}
        \centering
        \includegraphics[width=\linewidth,height=7cm]{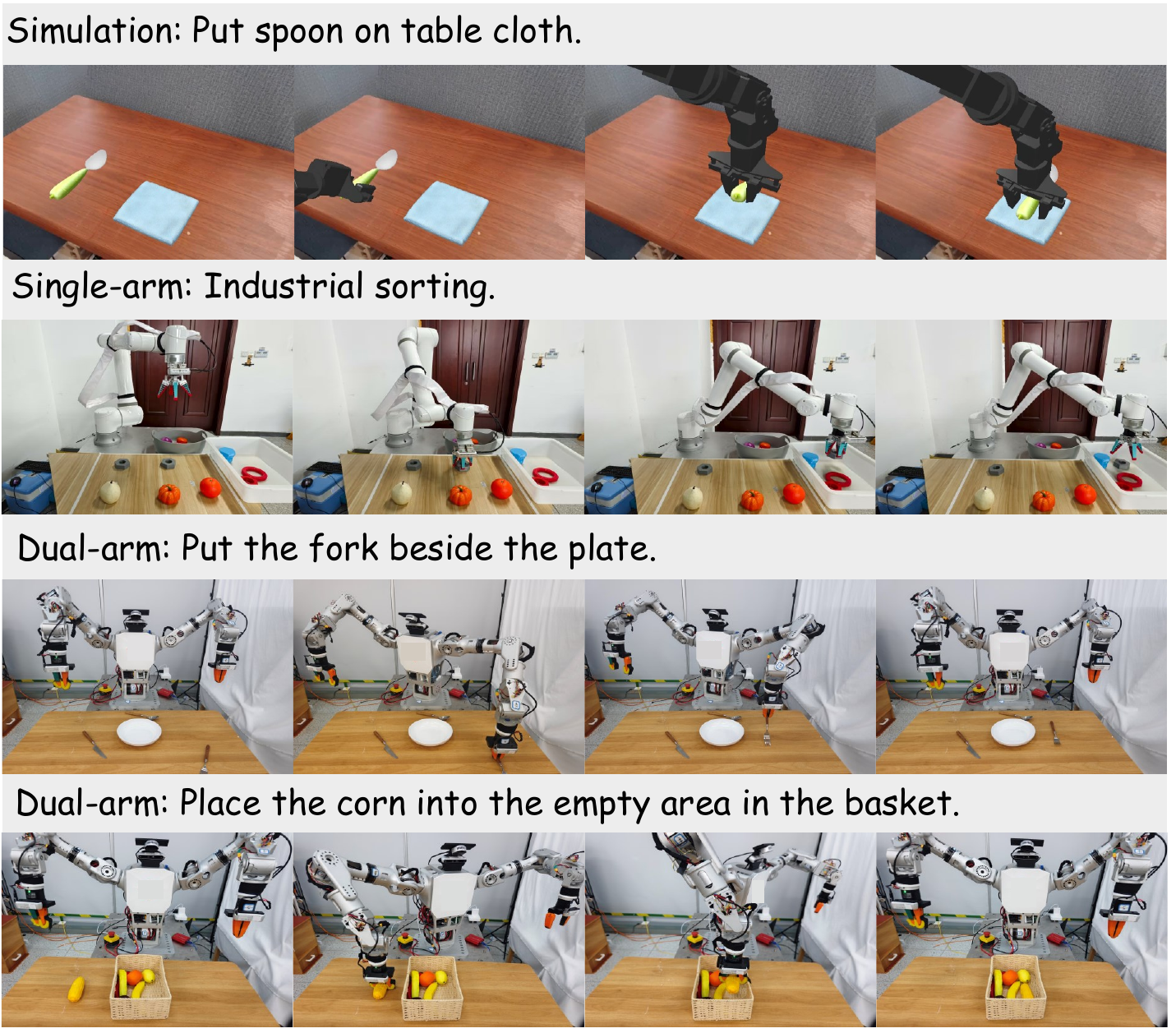}
        \caption{RoboMAP's zero-shot generalization across diverse environments (simulation, real-world) and robotic embodiments (simulated arm, single industrial arm, dual-arm humanoid).}
        \label{fig:realworld}
    \end{minipage}
    \hfill    
    \begin{minipage}[t]{0.49\textwidth}
        \centering
        \includegraphics[width=\linewidth,height=7cm]{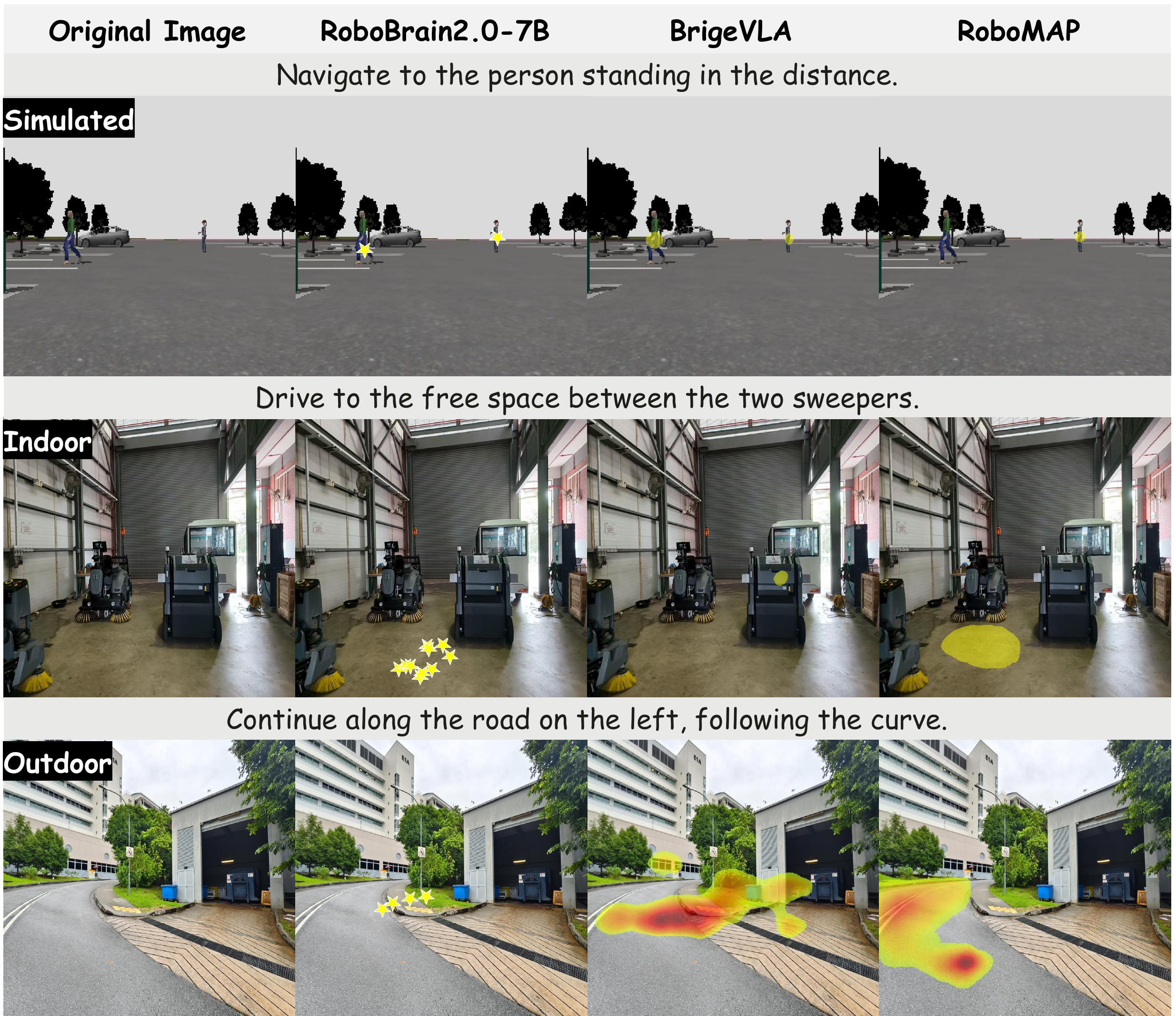}
        \caption{Zero-shot navigation tasks in three distinct environments. For clarity, the discrete point predictions are visualized as yellow stars and the heatmap colors have been adjusted for better visibility.}
        \label{fig:nav_visuals}
    \end{minipage}
\end{figure*}
\paragraph{Performance in the Real World}

\begin{table}[htbp]
\centering
\small
\caption{Zero-Shot Evaluation in SimplerEnv on WidowX Robot.}
\label{tab:simplerenv_widowx_succ}
\resizebox{\columnwidth}{!}{
    \begin{tabular}{l ccccc}
    \toprule
    \textbf{Model} & \textbf{Spoon$\rightarrow$Towel} & \textbf{Carrot$\rightarrow$Plate} & \textbf{Green$\rightarrow$Yellow} & \textbf{Eggplant$\rightarrow$Basket} & \textbf{Success Rate} \\
    \midrule
    \multicolumn{6}{l}{\textit{VLA-based Methods}} \\
    \midrule
    OpenVLA~\cite{kim2024openvla} & 0.0\% & 0.0\% & 0.0\% & 4.1\% & 1.0\% \\
    $\pi_0$ ~\cite{black2024pi0visionlanguageactionflowmodel} & 29.1\% & 0.0\% & 16.6\% &62.5\% & 27.1\% \\
    $\pi_0$ FAST~\cite{black2024pi0visionlanguageactionflowmodel} & 29.1\% & 21.9\% & 10.8\% & \underline{66.6\%} & 32.1\% \\
    \midrule
    \multicolumn{6}{l}{\textit{VLM-based Methods}} \\
    \midrule
    MOKA~\cite{liu2024moka} & 45.8\% & 41.6\% & 33.3\% & 12.5\% & 33.3\% \\
    SoFar~\cite{qi2025sofar} & 55.5\% & 56.9\% & \textbf{62.5\%} & 40.2\% & 53.8\% \\
    RoboPoint~\cite{yuan2024robopoint} & 16.7\% & 20.8\% & 8.3\% & 25.0\% & 17.7\% \\
    FSD~\cite{yuan2025seeing} & 41.6\% & 50.0\% & 33.3\% & 37.5\% & 40.6\% \\
    Embodied-R1~\cite{yuan2025embodied} & \underline{65.2\%} &\textbf{ 68.0\%} & 36.1\% &  58.3\% & \underline{56.2\%} \\
    \midrule
    \rowcolor{blue!10} \textbf{RoboMAP} & \textbf{66.0\%} & \underline{54.0\%} & \underline{52.0\%} & \textbf{70.0\%} & \textbf{60.5\%} \\
    \bottomrule
    \end{tabular}
}
\end{table}
\paragraph{Performance in the Real World}

To validate its utility, we tested RoboMAP in two real-world settings. First, on a custom dual-arm robot, we evaluated 5 tabletop manipulation tasks, as detailed in Table~\ref{tab:realworld_evaluation}. RoboMAP converts language commands into 2D grasp and place heatmaps, which guide a downstream pipeline using depth data and modules like SAM~\cite{kirillov2023segment} and GraspNet-1billion~\cite{fang2020graspnet} to compute final 6-DoF grasp and 3D place coordinates for execution by a motion planner. Figure~\ref{fig:realworld} shows several example executions.

RoboMAP achieves an \textbf{82\%} average success rate, significantly outperforming all baselines. This success stems from our dense heatmap representation. Unlike the sparse points from baselines, our heatmaps enable more precise object segmentation with SAM. The lower performance in \textit{Place [A] relative to [B]} is not a model grounding failure but a limitation of our post-processing, which selects only the peak-probability pixel. This heuristic fails on surfaces with height variations, even when the underlying heatmap correctly identifies the entire valid placement region.

Beyond accuracy, RoboMAP's inference speed of just \textbf{0.04 seconds} makes it highly practical for real-time applications. This is orders of magnitude faster than baselines like FSD (15.68s) and RoboBrain2.0-7B (8.07s), which incorporate lengthy iterative reasoning processes.

We plan to develop more sophisticated downstream policies to better leverage both the rich information in these heatmaps and high inference speed of our model.

In addition to the dual-arm setup, we further deployed RoboMAP on an \textbf{industrial single-arm robot} for high-precision sorting, such as locating a screw that is \textit{isolated from the main cluster}. The successful zero-shot experiments demonstrate that RoboMAP can produce a generalizable and embodiment-agnostic spatial representation. 
Videos of these robotic experiments will be released publicly.

\subsubsection{Embodied Performance on Mobile Navigation}

For \textbf{navigation} with the \textbf{mobile robot}, Figure~\ref{fig:nav_visuals} showcases RoboMAP's superior performance over the SOTA points-based Robobrain2.0-7B and BridgeVLA baselines across simulated, indoor, and outdoor environments. Our model accurately grounds complex instructions, such as interpreting qualifiers like \textit{in the distance} (Simulated), finding the \textit{free space between two sweepers} (Indoor), and \textit{following the curve} (Outdoor). In each environment, RoboMAP provides a precise navigational heatmap where the baselines are either inaccurate or fail completely.



\section{Conclusion, Limitation and Future Work}

We presented RoboMAP, a framework that grounds spatial language into \emph{adaptive affordance heatmaps}. By representing targets as distributions instead of points, our method captures spatial uncertainty to achieve state-of-the-art performance on key robotics benchmarks. Our method achieves an 82\% success rate in real-world manipulation, successfully generalizes to navigation, and operates at 25 Hz (0.04s per inference).

The primary limitation of our work lies not in the heatmap generation, but in how this rich information is underutilized by downstream policies. Our current heuristic of selecting only the single, peak-confidence pixel is overly simplistic, causing failures on surfaces with varying heights even when the heatmap correctly identifies the entire valid region. This directly motivates the need to design more sophisticated downstream policies.

Looking ahead, such policies could unlock two key research directions. First, they would enable smarter motion planning by leveraging the entire distribution to find optimal targets or recover from errors by targeting secondary peaks. Second, they would allow for uncertainty-aware behaviors, using the heatmap's confidence to trigger actions like proactively asking for clarification on ambiguous commands.

\bibliographystyle{IEEEtran}  
\bibliography{refs}           

\end{document}